\RequirePackage{fix-cm}
\documentclass[twocolumn]{svjour3}          
\smartqed  

\usepackage{amsmath,graphicx,url}

\usepackage{graphicx}
\usepackage{subfigure}
\usepackage{adjustbox}
\usepackage{multirow}
\usepackage{color}

\journalname{Machine Vision and Applications}
\begin{document}

\title{Detecting Personality and Emotion Traits in Crowds from Video Sequences\thanks{Thanks to Office of Naval Research Global (USA) and Brazilian agencies: CAPES, CNPQ and FAPERGS.}
}


\author{Rodolfo Migon Favaretto \and
        Paulo Knob \and
        Soraia Raupp Musse \and
        Felipe Vilanova \and
        \^Angelo Brandelli Costa
}


\institute{Rodolfo Migon Favaretto \and
        Paulo Knob \and
        Soraia Raupp Musse \at
              VHLab - Graduate Program in Computer Science - PUCRS, Av Ipiranga, 6681, Porto Alegre - RS - Brazil \\
              \email{rodolfo.favaretto@acad.pucrs.br}           
           \and
           Felipe Vilanova \and
        \^Angelo Brandelli Costa \at
              Graduate Program in Psychology - PUCRS, Av Ipiranga, 6681, Porto Alegre - RS - Brazil
}

\date{Received: date / Accepted: date}

\maketitle

\begin{abstract}
This paper presents a methodology to detect personality and basic emotion characteristics of crowds in video sequences. Firstly, individuals are detected and tracked, then groups are recognized and characterized. Such information is then mapped to OCEAN dimensions, used to find out personality and emotion in videos, based on OCC emotion models. Although it is a clear challenge to validate our results with real life experiments, we evaluate our method with the available literature information regarding OCEAN values of different Countries and also emergent Personal distance among people. Hence, such analysis refer to cultural differences of each country too. Our results indicate that this model generates coherent information when compared to data provided in available literature, as shown in qualitative and quantitative results.
\keywords{Computer vision \and crowd features \and Big-five model \and cultural dimensions \and crowd emotion}
\end{abstract}
 
\section{Introduction}
\label{sec:intro}

Crowd analysis is a phenomenon of great interest in current applications. Surveillance, entertainment and social sciences are examples of fields that can benefit from the development of this area of study. Literature presents different applications of crowd analysis, like counting people in crowds \cite{Chan2009,cai2014}, group and crowd movement and formation~\cite{Solmaz2012,Zhou2014,Ricky:15,jo2013review} and detection of social groups in crowds~\cite{solera_2013,Shao2014,Feng2015,Chandran2015}. Normally, these approaches are based on personal tracking or optical flow algorithms, and handle with features like walking speed, directions and distances over time. Specifically on this subject, one study investigated cultural difference in videos from different countries: Chattaraj et al.~\cite{CHATTARAJ2009} suggested that cultural and population differences could produce deviations in speed, density and flow of the crowd.

In this paper, we propose to detect personality aspects based on the Big-five personality model, also referenced as (OCEAN) \cite{costa07}, using individuals behaviors automatically detected in video sequences. For this, we used the NEO PI-R~\cite{costa07} which is the standard questionnaire measure of the Five Factor Model. Also, we use such psychological traits to identify some primordial emotions in the crowd. Using a similar mapping as proposed by Saifi et al.~\cite{saifi2016approach}, we were able to identify the level of some emotions for each individual, like happiness or fear, according its OCEAN level.

This paper is organized as follows: Section~\ref{sec:related} presents related works on personality and crowd emotion detection. Section~\ref{sec:model} describes our model to detect personality traits for individuals and groups. Moreover, we use this psychological traits to identify different levels of emotions in the crowd, like if an agent is happy or angry, in Section~\ref{sec:emotionDiscovery}.
Experimental results are addressed in Section~\ref{sec:results}, where we deal with the challenge to evaluate our approach with real life. Conclusions and future work are presented in Section~\ref{sec:conclusions}.

\section{Related Work}
\label{sec:related}

This section discusses some topics concerned with personality and emotion detection in crowds domain.

Personality may be labeled as deep psychological individual level trait~\cite{cattell50}. Trait is an inference made after observed behaviors that seeks to explain its regularity ~\cite{hall98}. Raymond Cattel is commonly referred as the one who developed the methodology which permitted the objective grouping of hundreds of trait descriptors in a set of higher level factors~\cite{digman90}. Cattell~\cite{cattell48} developed a taxonomy of individual differences that consisted of 16 primary factors and 8 second-order factors. Nevertheless, attempts to replicate his work were unsuccessful~\cite{fiske48} and researchers agreed that only the 5-factor model matched his data, originating the Big Five personality model.

Nowadays, researchers agree that there are five robust orthogonal traits which effectively matched personality attributes~\cite{digman90}, known as the Big Five: Openness to experience (“the active seeking and appreciation of new experiences”); Conscientiousness (“degree of organization, persistence, control and motivation in goal directed behavior”); Extraversion (“quantity and intensity of energy directed outwards in the social world”); Agreeableness (“the kinds of interaction an individual prefers from compassion to tough mindedness”); Neuroticism (how much prone to psychological distress the individual is)~\cite{lordw07}.

The NEO PI-R~\cite{costa92} is one of the most used instrument based on the Big Five personality theory. It assesses the normal adult personality and is internationally recognized as a gold standard for personality assessment. One of its advantages is that it further specifies six facets within each personality trait and have data from several countries which easily allows cross-cultural comparisons~\cite{McCrae2002,McCrae05}. Although the empirical evidence matching individual level traits 
and crowd behavior is not strong (one of the few examples is~\cite{Barry97}), the Big-Five personality model is widely used to model computational crowd simulation~\cite{kaup06,Durupinar08,Guy11}. In general, in such methods, OCEAN model allows to simulate a crowd with individual level parameters based on the expected behaviors of the agents.

\begin{figure*}[t]
\centering
\includegraphics[scale=0.68]{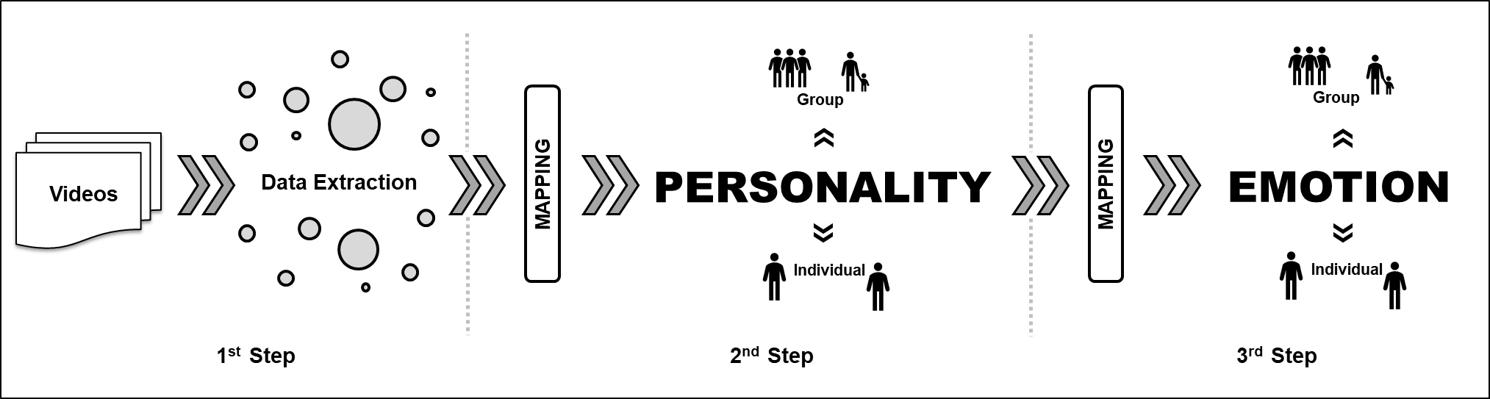}
\caption{Three main steps of our approach: ($1^{st}$) individual tracking and data extraction, ($2^{nd}$) individual data is mapped to individual and group personality (OCEAN) traits and ($3^{rd}$) personality traits are mapped to individual and group basic emotions. These steps are detailed in Figure \ref{fig:flow_detailed}.}
\label{fig:approach}
\end{figure*}

Several models have been developed to explain and quantify basic emotions in humans and other animals. One of the most cited is the model proposed by Paul Ekman~\cite{ekman1971constants} which considers the existence of 6 universal emotions based on cross-cultural facial expressions (anger, disgust, fear, happiness, sadness and surprise). Other approaches such as Affective Neuroscience postulate, from an evolutionary perspective, consider other groups of emotions such as fear, rage/anger and sadness/panic~\cite{montag2017primary}.

Panksepp’s theory claims that individual differences in primary emotional systems may represent the phylogenetically oldest parts of human personality. In order to support this idea, the author links individual differences in primary emotions and the big five model of personality. for example, seeking is robustly linked with openness, high play with higher extraversion, high care and low anger are associated with higher agreeableness, high scores on fear, sadness and rage with high neuroticism~\cite{davis2003affective,davis2011brain}. Finally, other models such as the proposed by Ortony, Clore, and Collins (commonly referred to as the OCC model) distinguish 22 emotion types manly bases on its expression. This is the model used in this paper~\cite{clore2013psychological}.

In the work proposed by Gorbova and collaborators~\cite{Gorbova2017}, authors present a system of automatic personality screening from video presentations in order to make a decision whether a person has to be invited to a job interview. The automatic personality screening is based on visual, audio and lexical cues from short video-clips. The system is build to predict candidate scores of 5 Big Personality Traits and to estimate a final decision, to which degree the person from video-clip has to be invited to the job interview.

Baig et al.~\cite{baig2015perception} focus their work on the perception of emotion due to crowd behavior. For this end, they propose an approach based on probabilistic modeling, which is trained to perceive the emotions of people in a given area. There, it uses camera sensors in order to track the motion of the individuals in the crowd. They use data mining techniques to identify different behaviors and events. Agents can experience different types of interactions during the simulation, which can arouse negative or positive emotions. Tests showed that the algorithms for emotions detection performed well with all tested scenarios and types of interaction. Yet, they comment that it is still necessary to include more complex interactions and motivations in a more complex probabilistic model in order to have a more realistic model.

Following the emotion detection in crowds subject, Rabiee et al.~\cite{rabiee2016emotion} mention that emotions can be valuable traits in the understanding of the crowd. Moreover, they say that the majority of methods proposed are just based on low-level visual features, leaving a huge semantic gap between these low-level features and high-level concept of crowd behavior. Therefore, in their work, it is proposed an attribute-based strategy to generate a crowd dataset with both annotations of abnormal crowd behavior and crowd emotion. In fact, the dataset is the main contribution of their work, which can be used as a benchmark for computer vision as well to help to understand the correlation between crowd behavior and emotion.

Saifi et al.~\cite{saifi2016approach} proposed a mapping between psychological traits and emotions. In their work, they aim to model emergent emotions and behaviors in a simulated crowd, based on the personality of each individual. For this, the OCEAN~\cite{costa07} personality model is combined with the OCC~\cite{ortony1990cognitive} emotional model in order to find the susceptibility of each of the five OCEAN personality factors to feel each OCC emotion. They use fuzzy logic to model the critical emotions when in presence of an unexpected event, which can trigger a specific behavior. In their method, it is possible to observe a crowd and the effect of events, like how calm or nervous agents react to it. Moreover, the user is able to predict future situations or events, which can lead him/her to make better decisions at the right time.

Following a similar way, Zhang et al.~\cite{zhang2017exploring} proposed a novel crowd representation named ``crowd mood", which should be based on the spacing interactions of the individuals and the structural levels of motion patterns in crowds. They affirm that ``Basic types of crowd motion can reflect representative
emotions of the crowd.". Therefore, the main idea is to project varied types of statistical motion features and, based on this, assign emotional labels to the behavior of the crowd. The preliminary achieved results  showed an interesting performance
in the given tasks, offering a promising tool which represents crowd behaviors semantics and can be applied to varied crowd tasks.

In this paper, the idea is to map parameters from individual and group behaviors automatically detected from video sequences of different countries, to OCEAN dimensions. We extended a previous work~\cite{favaretto:sib2017} in order to detect groups and find their OCEAN levels as well. Then, we use OCEAN of groups and individuals to find out their emotion in the video sequence, following a mapping similar with the one proposed by Saifi et al.~\cite{saifi2016approach}. While their work present a model to simulate crowds with psychological traits and emotions, our work focus on identifying such traits in video sequences. Further details are going to be presented in Section~\ref{sec:model}. In this sense, our contribution is a model based on a set of equations that handle the individual parameters obtained from videos and map  them to the methodology that compose the Big-Five personality model for individuals and groups, proposing a mapping function to find out emotion values. Our method does not require any training or specific dataset since we based our prototype on the theories behind the OCEAN and OCC (emotion model).

\section{The proposed approach}
\label{sec:model}

\begin{figure}[h]
\centering
\includegraphics[scale=0.65]{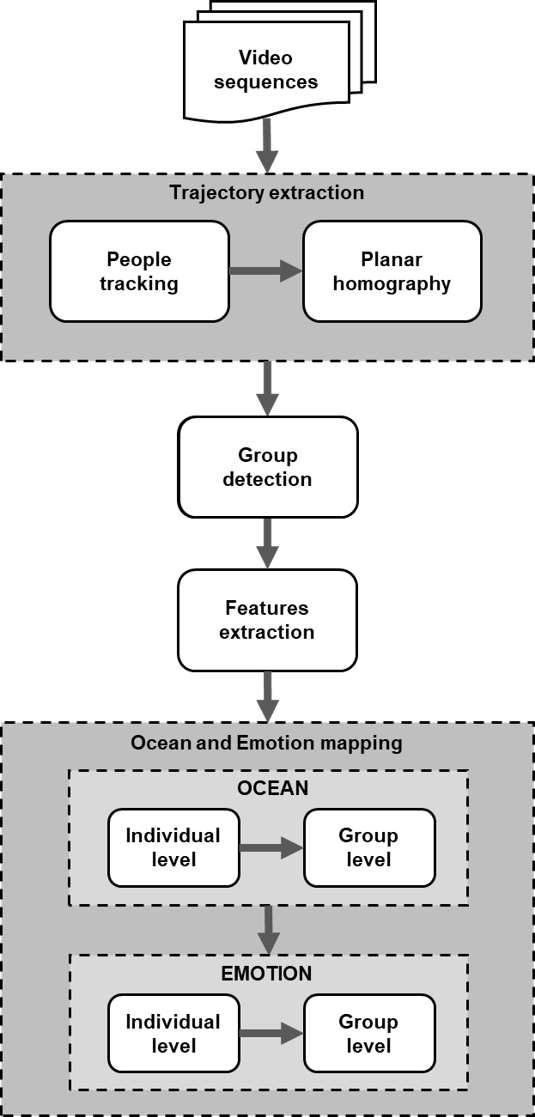}
\caption{Flow chart of our approach: Initially, people trajectories are extracted and passed by a homography transformation process. Based on these trajectories, groups are detected and its features are extracted. These features are the inputs for the OCEAN and EMOTION acquisition process.}
\label{fig:flow_detailed}
\end{figure}

Our model presents three main steps responsible for following: video data extraction, personality and emotion analysis. These steps are illustrated in the overview of the method in Figure~\ref{fig:approach}. The first step aims to obtain the individual trajectories from observed pedestrians in real videos. Using these trajectories, we detect groups and extract data which are useful for second step, that is responsible for personality analysis of groups and individuals. Once we have concluded the second step, we have enough information to follow with the third step, which consists of emotion detection of individuals and groups according to OCEAN values. Figure~\ref{fig:flow_detailed} illustrates a more detailed flow chart of our approach and Section~\ref{sec:step1} presents further details.

\subsection{Individuals and Groups Data Extraction}
\label{sec:step1}

\begin{figure}[h]
\centering
\subfigure[fig:N_15][Trajectories from tracking process]{\includegraphics[width=3in]{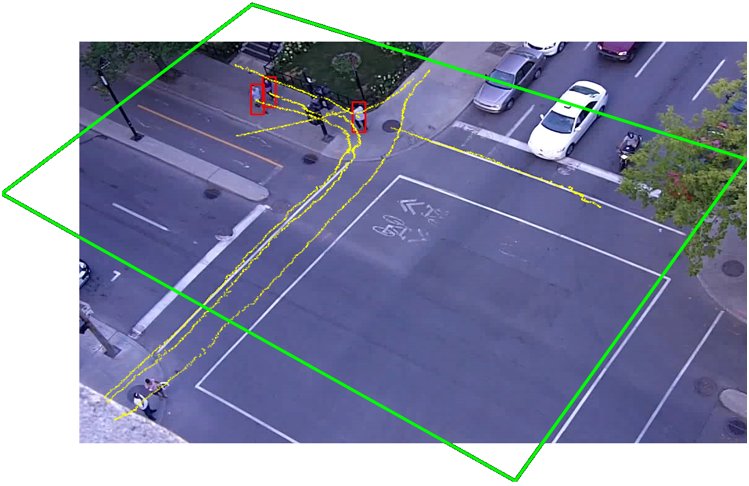}}
\subfigure[fig:N_34][Trajectories after the homography transformation]{\includegraphics[width=3in]{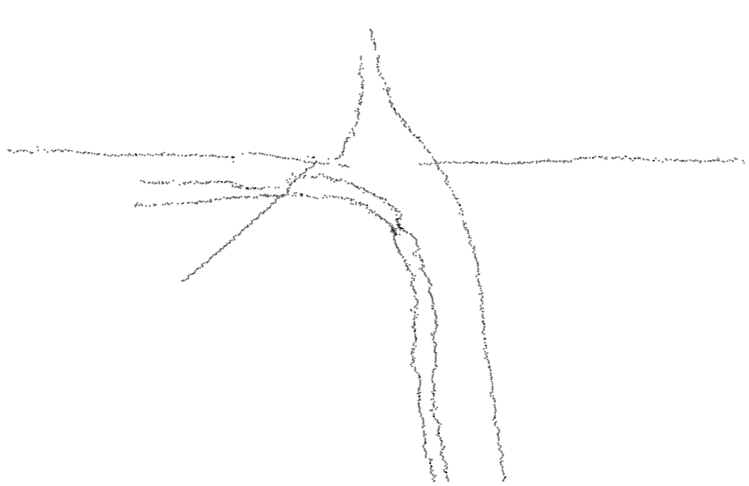}}
\caption{Homography transformation process: (a) people trajectories (yellow points in the scene) are obtained from the tracker process and rectified by homography transformation process (b).}
\label{fig_homography}
\end{figure}

Initially, the information about people from real videos is obtained using a tracker~\cite{Bins2013} to recover people trajectories. In order to transform image in world coordinates, we assume that the head position is on the ground place ($z=0$), and then we used a \textit{planar homography} to rectify a perspective image and generate world coordinates. In computer vision, \textit{planar homography} is defined as a projective mapping from one plane to other. Then we use the homography  to rectify a perspective image, in this case, to generate a ``plan" view of trajectories from a ``perspective" photo. Figure~\ref{fig_homography} illustrates the mapping of the trajectories points from  image coordinates to a orthogonal plan view (world coordinates). The coordinates of trajectories are used to calculate the motion parameters (speed of travel, direction) for each agent.

We compute information for each person $i$ at each timestep: \textit{i)} 2D position $\vec{x}_i$ (meters); \textit{ii)} speed $s_i$ (meters/frame); \textit{iii)} angular variation $\alpha_i$ (degrees) w.r.t. a reference vector $\vec{r}=(1,0)$; \textit{iv)} isolation level $\varphi_i$; \textit{v)} socialization level $\vartheta_i$; and \textit{vi)} collectivity $\phi_i$. The isolation level $\varphi$ is computed as shown in Equation \ref{eq:isolation}:

\begin{equation}
  \varphi_i = \left\{
  \begin{array}{ll}
  1, & \text{~if~} n_{social} = 0 \\
  \frac{\frac{1}{n_{social}}\sum_{j=0}^{n_{social}-1}d(\vec{x}_i,\vec{x}_j)}{d_{Hall}}, & \text{~otherwise}
  \end{array},
  \label{eq:isolation}
  \right .
\end{equation}

\noindent where $n_{social}$ is the number of individuals in the social space\footnote{Social space is related to $3.6$m~\cite{hall98}.} according to Hall's proxemics~\cite{hall98}, $d$ is a simple function to calculate the Euclidean distance between two individuals $i$ and $j$ ($\vec{x}_i$ and $\vec{x}_j$ are, respectively, the positions of individuals $i$ and $j$) and $d_{Hall} = 3.6$ is the distance (in meters) around an individual that represents its personal social space. 
The socialization level $\vartheta_i$ of an individual $i$ is calculated according to the Equation~\ref{eq:socialization}:

\begin{equation}
\vartheta_i = \left\{
\begin{array}{ll}
0, & \text{~if~} n_{social} = 0 \\
\frac{n_{social}}{\rho}, & \text{~otherwise}
\end{array},
\label{eq:socialization}
\right .
\end{equation}

\noindent where $\rho$ is the total number of individuals in the analyzed frame.
To compute the collectivity affecting individual $i$ and exerted by $n_{social}$ individuals in social space (as presented in~\cite{Favaretto:Sib:2016}), we use the Equation~\ref{eq:collectivity}, as follow:

\begin{equation}
\phi_i = \sum_{j=0}^{n-1} \gamma e^{(-\beta \varpi(i,j)^{2})},
\label{eq:collectivity}
\end{equation}

\noindent where the collectivity between two individuals $i$ and $j$ is calculated as a decay function of $\varpi(i,j) = s(s_i,s_j).w_1+o(\alpha_i,\alpha_j).w_2$, considering $s$ and $o$ respectively their speed and orientation differences, 
and $w_1$ and $w_2$ are constants that should regulate the offset in meters and radians. We have used $w_1=1$ and $w_2=1$. $\gamma = 1$ is the maximum collectivity value when $\varpi(i,j)=0$, and $\beta = 0.3$ is empirically defined as decay constant. Hence, $\phi_{i}$ is a value in the interval $[0;1]$. Therefore, for each individual $i$ at each frame $f$ in a video sequence, we compute vector $\vec{V_{i,f}}$ of extracted data where $\vec{V_{i,f}} = \left [s_{i,f}, \alpha_{i,f}, \varphi_{i,f}, \vartheta_{i,f}, \phi_{i,f} \right ]$. Yet, the average values for all frames in the video sequence are represented through the vector $\vec{V_i}$ of individual $i$ where $\vec{V_i} = \left [\bar{s}_i, \bar{\alpha}_i, \bar{\varphi}_i, \bar{\vartheta}_i, \bar{\phi}_i \right ]$.

For groups detection, we use the computed parameters $s$, $o$ and $d$ which respectively state for speed and orientation variation and distance of each pair of agents $i$ and $j$. We use the notion of distances based on the \textit{proxemics} described by Hall~\cite{Hall:1990} to define that two agents belong to the same group according with three tests empirically defined: If $(d(x_i,x_j) <= 1.2 \text{meter})$ and $(o(\alpha_i,\alpha_j) <= 15^{\circ})$ and ($s(s_i,s_j) < \beta$), where $\beta=5\%$ of higher speed. Based on this set of rules, agents are grouped in pairs. In a next step, we check which pairs have one individual in common, and merge them into larger groups. This process is performed until the group formation does not share individuals, i.e. they are disjoint. For each group $g$ detected in video sequence, we define a vector $\vec{G_g} = \left [n_g, \vec{I}_g, \bar{s}_g, \bar{\alpha}_g, \bar{d}_g \right ]$ of extracted data, where $n_g$ is the number of individuals in $g$, $\vec{I}_g$ states for a vector containing the id of each individual in $g$ and $\bar{s}_g$, $\bar{\alpha}_g$ and $\bar{d}_g$ present the average values of speed, orientation and distance applied by individuals in group $g$ during the video sequence.

At the end of this step, we have $\vec{V}$ for all individuals per frame and averaged for the video sequence and $\vec{G}$ for all groups averaged for the video sequence. In next section we describe how these data is used to find out the personality traits in our work.

\subsection{Mapping crowd features in OCEAN Dimensions}

We used the empirically equations proposed by Favaretto et al.~\cite{favaretto:sib2017} in a previous work which goal is to map individual and group characteristics in OCEAN cultural dimensions. Basically, the method proposes to ``answer" 25 items from NEO PI-R inventory for each individual in the video sequence. Although the complete version of NEO PI-R has $240$ items, 25 of them were selected since they have a direct relationship with crowd behavior. Each question is mapped to an equation using only information contained in $V$ for each $i$. For example, in order to represent the item ``1 - Have clear goals, work to them in orderly way'', we consider that the individual $i$ should have a high velocity $s$ and low angular variation $\alpha$ to have answer compatible with ``Strong". So the equation for this item is $Q_1 = s_i+\frac{1}{\alpha_i}$. Table \ref{tab:equations} shows the equations used for each considered question.

\begin{table}[h]
   \renewcommand{\arraystretch}{1.45}
   \centering
   \scriptsize
   \caption{Equations from each NEO PI-R selected item.}
   \begin{adjustbox}{max width=8.4cm}
     \begin{tabular}{l|c}
        \hline\noalign{\smallskip}
        NEO PI-R Item & Equation \\
        \noalign{\smallskip}
        \hline
        \noalign{\smallskip}
        \hline
        \noalign{\smallskip}
        1 - Have clear goals, work to them in orderly way & $Q_1 = s_i+\frac{1}{\alpha_i}$ \\
         \noalign{\smallskip}
        \hline
        \noalign{\smallskip}
        2. Follow same route when go somewhere & $Q_2 = \alpha_i$ \\
        \noalign{\smallskip}
        \hline
        \noalign{\smallskip}
        3. Shy away from crowds & \multirow{6}{*}{$Q_{3-8} = \varphi_i$} \\
        4. Don’t get much pleasure chatting with people & \\
        5. Usually prefer to do things alone & \\
        6. Prefer jobs that let me work alone, unbothered & \\
        7. Wouldn’t enjoy holiday in Las Vegas & \\
        8. Many think of me as somewhat cold, distant & \\
        \noalign{\smallskip}
        \hline
        \noalign{\smallskip}
        9. Rather cooperate with others than compete & \multirow{2}{*}{$Q_{9-10}=\phi_i$} \\
        10. Try to be courteous to everyone I meet & \\
        \noalign{\smallskip}
        \hline
        \noalign{\smallskip}
        11. Social gatherings usually bore me & $Q_{11}=\varphi_i + std(\alpha_i)$ \\
        \noalign{\smallskip}
        \hline
        \noalign{\smallskip}
        12. Usually seem in hurry & $Q_{12}=s_i+\alpha_i$  \\
        \noalign{\smallskip}
        \hline
        \noalign{\smallskip}
        13. Often disgusted with people I have to deal with & $Q_{13}=\varphi_i+\frac{1}{\phi_i}$ \\
        \noalign{\smallskip}
        \hline
        \noalign{\smallskip}
        14. Have often been leader of groups belonged to & $Q_{14}=\phi_i+\vartheta_i+\frac{1}{\alpha_i}$  \\
        \noalign{\smallskip}
        \hline
        \noalign{\smallskip}
        15. Would rather go my own way than be a leader & $Q_{15}=\frac{1}{Q_{14}}$ \\
        \noalign{\smallskip}
        \hline
        \noalign{\smallskip}
        16. Like to have lots of people around me & \multirow{6}{*}{$Q_{16-21}=\vartheta_i$} \\
        17. Enjoy parties with lots of people & \\
        18. Like being part of crowd at sporting events & \\
        19. Would rather a popular beach than isolated cabin & \\
        20. Really enjoy talking to people & \\
        21. Like to be where action is & \\
        \noalign{\smallskip}
        \hline
        \noalign{\smallskip}
        22. Feel need for other people if by myself for long & \multirow{4}{*}{$Q_{22-25}=\vartheta_i+\phi_i$} \\
        23. Find it easy to smile, be outgoing with strangers & \\
        24. Rarely feel lonely or blue & \\
        25. Seldom feel self-conscious around people & \\
        \noalign{\smallskip}
        \hline
     \end{tabular}
   \end{adjustbox}
   \label{tab:equations}
\end{table}

Once all questions $k$ (in the interval $[1;25]$) have been answered for all individuals $i$, we have $\vec{Q_{i,k}^f}$ for each frame $f$. In addition, we computed the average values to have one vector $\vec{Q_{i,k}}$ per video. According to NEO PI-R definition, each of the questions $\vec{Q'_{k}}$ are associated to one of the Big Five dimensions and some questions should invert the values, because an item score 4 (Strongly Agree) can represent a high or low value of a certain personality trait. So, to get the correct values, we applied a factor to the questions which score should be inverted: $\vec{Q^*_{i,k}} = 4 - \vec{Q'_{i,k}}$, as shown in next equations:

\begin{equation}
O_i = \frac{Q^*_{i,2}}{\varrho},
\end{equation}
\begin{equation}
C_i = \frac{Q'_{i,1}}{\varrho},
\end{equation}
\begin{equation}
E'_i = Q'_{i,3} + Q'_{i,12} + Q'_{i,14} + \sum_{q=16}^{23} Q'_{i,q},
\label{eq:e1}
\end{equation}
\begin{equation}
E^*_i = \sum_{q=4}^{8} Q^*_{i,q} + Q^*_{i,11} +  Q^*_{i,15},
\label{eq:e2}
\end{equation}
\begin{equation}
E_i = \frac{(E'_i+ E^*_i)}{\varrho},
\label{eq:e3}
\end{equation}
\begin{equation}
A_i = \frac{\sum_{q=9}^{10}Q'_{i,q}}{\varrho},
\end{equation}
\begin{equation}
N_i = \frac{Q'_{i,13}+ \sum_{q=24}^{25}Q^*_{i,q}}{\varrho},
\end{equation}

where $\varrho$ represents the percentage of questions from the total, in each dimension (O, C, E, A and N), respectively 4\%, 4\%, 72\%, 8\% and 12\%.

\subsection{Emotion Detection}
\label{sec:emotionDiscovery}

In this paper we connect personality and emotion traits in order to detect data in the individuals in video sequences.
As mentioned by~\cite{revelle2009personality}:

"A helpful analogy is to consider that personality is to emotion as climate is to weather. That is, what one expects is personality, what one observes at any particular moment is emotion." That is the reason why, in our work, we correlated emotion traits with visual behaviors that will be perceived in pedestrian motion.

Our work proposes an emotion mapping based on personality traits (i.e. OCEAN) found for each individual present in the video sequence. First, we selected four emotions from OCC~\cite{ortony1990cognitive} model: Fear, Happiness, Sadness and Anger. It is important to notice that we chose only four emotions that, in our opinion, are the most visible when relating with motion behavior, which is our case in video sequences with pedestrians. Any other from the total of 22 emotions proposed in OCC model could also be mapped.

Once we have computed the personality traits for an individual, we propose a way to map these personality traits into the four considered emotions. In addition to vector $\vec{V}_{i,f}$ for individual $i$ per frame $f$, we included personality and emotion parameters as follows: $\vec{V}_{i,f} = \left [s_{i,f}, \alpha_{i,f}, \varphi_{i,f}, \vartheta_{i,f}, \phi_{i,f}, \vec{P}_{i,f}, \vec{E}_{i,f} \right ]$. $\vec{P}_{i,f}$ states for values of OCEAN personality $\left [O_{i,f}, C_{i,f}, E_{i,f}, A_{i,f}, N_{i,f},\right ]$, while $\vec{E}_{i,f}$ states for values of emotion based on OCC model: $\left [F_{i,f}, H_{i,f}, S_{i,f}, An_{i,f}\right ]$. Values of $\vec{P}$ are in the interval $[0;1]$ while values of $\vec{E}$ are from $[-3;3]$. For groups analysis, we only considered parameters of personality and emotion in the detected groups $g$ in the video sequence. In addition to previously defined $\vec{G_g} = \left [n_g, \vec{I}_g, \bar{s}_g, \bar{\alpha}_g, \bar{d}_g \right ]$, we include $\vec{P}_g$ and $\vec{E}_g$ containing the average values of $n_g$ individuals in $g$. 

In order to map from OCEAN to emotion parameters we observe some aspects in literature ~\cite{costa07}:

\begin{itemize}
\item  O- : person is close to interact with others;
\item  O+ : person is aware of his/her feelings;
\item  C+ : person is optimistic;
\item  C- : person is pessimist;
\item E+ : person has a strong relationship with positive emotions;
\item E- : person presents relationship with negative emotions;
\item A+ : person has a strong relationship with positive reactions;
\item A- : person presents relationship with negative reactions;
\item N-: known by the emotional stability;
\item N+ : person feels negative emotions;
\end{itemize}

Such data resulted in empirical definitions included in Table~\ref{tab:emotionMapping}, that shows the mapping from OCEAN traits to the chosen emotions. In fact, we were not the first one to propose this type of mapping. Saifi et al.~\cite{saifi2016approach} proposes similar data for a different approach where authors were interested in providing critical emotions in crowd simulators. Davis and Panksepp \cite{davis2011brain} also proposed a similar approach unifying basic emotions with personality. 

In Table~\ref{tab:emotionMapping}, the plus/minus signals along each factor represent the positive/negative value of each one. For example, O+ stands for positive values (i.e. O $\geq$ 0.5) and O- stands for negative values (i.e. O $<$ 0.5)). A positive value for a given factor (i.e. 1) means the stronger the OCEAN trait is, the stronger is the emotion too. A negative value (i.e. -1) does the opposite, therefore, the stronger the factor's value, the weaker is a given emotion. A zero value means that a given emotion is not affected at all by the given factor. To better illustrate, a hypothetical example is given: if an individual has a high value for Extraversion (for example, E = 0.9), following the mapping in Table~\ref{tab:emotionMapping}, this individual can present signals of happiness (i.e. If E+ then Happiness= 1) and should not be angry (i.e. If E+ then Anger= -1).

\begin{table}[!htb]
   \renewcommand{\arraystretch}{1}
   \centering \small
   \caption{Emotion mapping from OCEAN to OCC. The plus/minus signals along each factor represent the positive/negative value of each one.}
   \begin{adjustbox}{max width=\textwidth}
     \begin{tabular}{ccccc}
     \hline\noalign{\smallskip}
     \textbf{Factor} & \textbf{Fear} & \textbf{Happiness} & \textbf{Sadness} & \textbf{Anger} \\
     \noalign{\smallskip}
        \hline
        \hline
        \noalign{\smallskip}
     O+ & 0 & 0 & 0 & -1 \\
     \noalign{\smallskip}
        \noalign{\smallskip}
     O- & 0 & 0 & 0 & 1 \\
     \noalign{\smallskip}
        \noalign{\smallskip}
     C+ & -1 & 0 & 0 & 0 \\ 
     \noalign{\smallskip}
        \noalign{\smallskip}
     C- & 1 & 0 & 0 & 0 \\ 
     \noalign{\smallskip}
        \noalign{\smallskip}
     E+ & -1 & 1 & -1 & -1 \\ 
     \noalign{\smallskip}
        \noalign{\smallskip}
     E- & 1 & 0 & 0 & 0 \\ 
     \noalign{\smallskip}
        \noalign{\smallskip}
     A+ & 0 & 0 & 0 & -1 \\ 
     \noalign{\smallskip}
        \noalign{\smallskip}
     A- & 0 & 0 & 0 & 1 \\ 
     \noalign{\smallskip}
        \noalign{\smallskip}
     N+ & 1 & -1 & 1 & 1 \\ 
     \noalign{\smallskip}
        \noalign{\smallskip}
     N- & -1 & 1 & -1 & -1 \\ 
     \noalign{\smallskip}
        \hline
     \end{tabular}
   \end{adjustbox}
   \label{tab:emotionMapping}
\end{table}

Finally, emotions are normalized in the interval $[0;1]$ considering the lowest and highest achieved values in the video, in order to keep the obtained scale.

\section{Experimental Results}
\label{sec:results}

In this section we discuss some results obtained with our approach. We organized it into four different analysis: \textit{i)} Ocean and Emotion recognition in spontaneous videos (Sections \ref{sec:ocean_analysis} and \ref{sec:emotion_analysis}, respectively), \textit{ii)} an Ocean and Emotion comparison with Literature (Section \ref{sec:compara}), \textit{iii)} an Ocean and Emotion analysis in the Fundamental Diagram experiment (Section \ref{sec:ocean_emotion_fd})  and \textit{iv)} a quantitative analysis of personal space (Section \ref{sec:personal_space}).

\subsection{OCEAN analysis}
\label{sec:ocean_analysis}

\begin{figure}[h]
\centering
\includegraphics[scale=0.325]{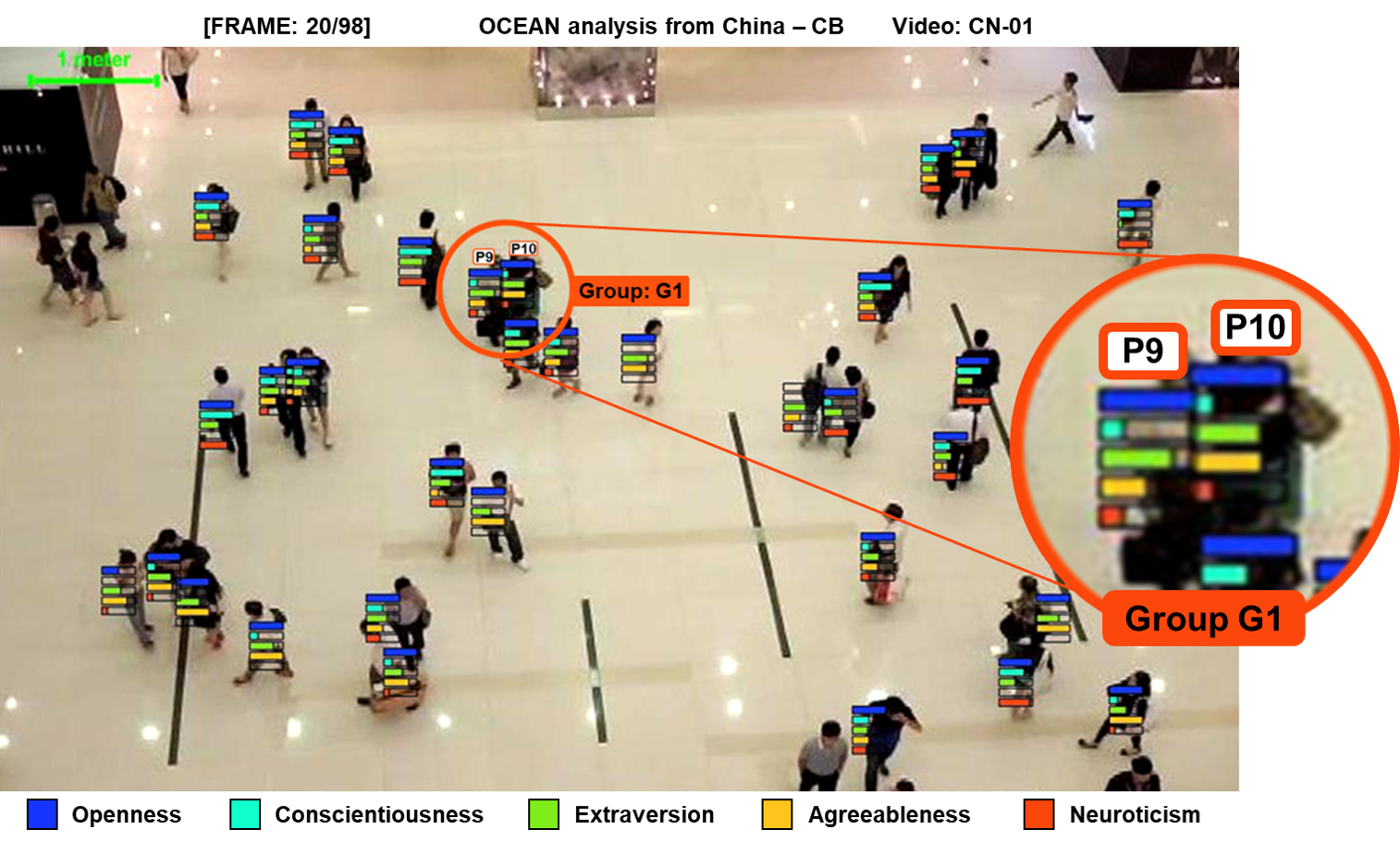}
\caption{OCEAN representation of each individual in the scene. Each color bar is related to one OCEAN dimension: blue is \textit{Openness}, cyan is \textit{Conscientiousness}, green is \textit{Extraversion}, yellow is Agreeableness and red, \textit{Neuroticism}. Information about highlighted group $G1$ (formed by the individuals $P9$ and $P10$) is detailed in Figure~\ref{fig:ocean_groups}.}
\label{fig:ocean_rep}
\end{figure}

In this section, we present OCEAN analysis involving spontaneous videos (crowds  in public spaces). Firstly, we calculate the OCEAN of each individual in the video at each frame. Once we get the individuals OCEAN, the group OCEAN is computed  by the average of the individuals' OCEAN that are part of the group. Figure \ref{fig:ocean_rep} shows a representation of each individual OCEAN in a determined frame. We used five color box that represent the five dimensions, where blue is related to Openness, cyan indicates Conscientiousness, green indicates Extraversion, yellow means Agreeableness and red, Neuroticism.

\begin{figure}[h]
\centering
\subfigure[fig:g1][Group $G1$]{\includegraphics[width=3.27in]{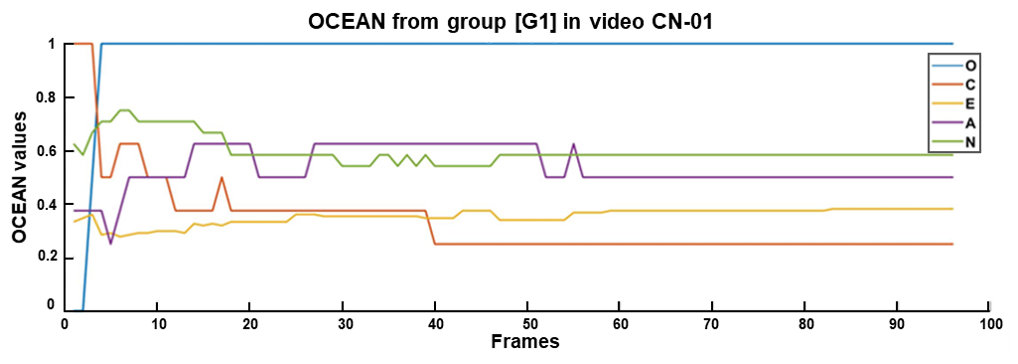}}
\subfigure[fig:g1_9][Individual $P9$]{\includegraphics[width=3.27in]{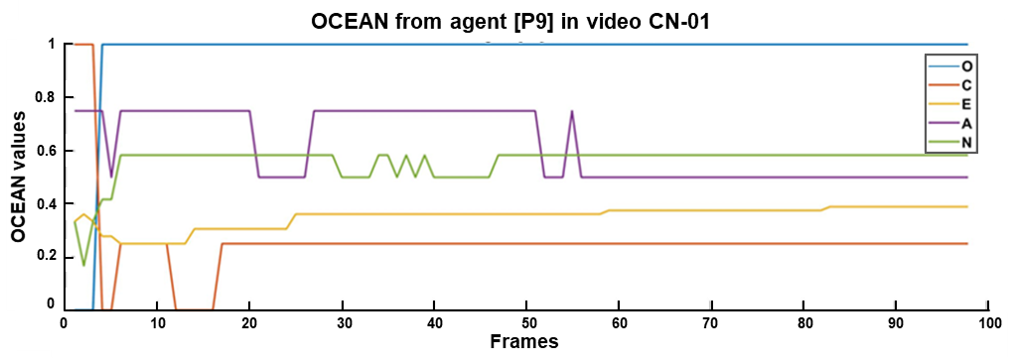}}
\subfigure[fig:g1_10][Individual $P10$]{\includegraphics[width=3.27in]{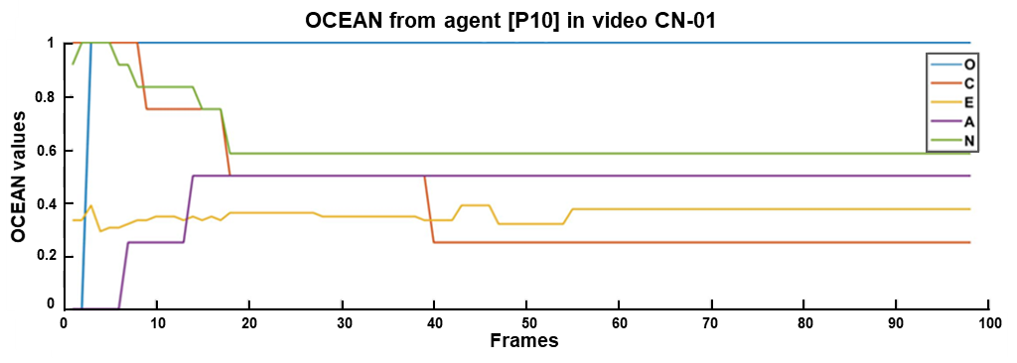}}
\caption{OCEAN data over the time observed in the group of individuals highlighted in Figure \ref{fig:ocean_rep}: data from group $G1$ (a) and its individuals $P9$ (b) and $P10$ (c).}
\label{fig:ocean_groups}
\end{figure}

The group of individuals highlighted in Figure~\ref{fig:ocean_rep} is composed of two people. The OCEAN of this group (named $G1$) over the time is illustrated in Figure~\ref{fig:ocean_groups}(a), and obtained by the average OCEAN of its individuals ($P9$ and $P10$), presented in Figures~\ref{fig:ocean_groups}~(b) and (c). As can be seen in such figures, the two individuals present more motion variation at the beginning of the video and then they keep the same motion characteristics until the end of the short movie (which duration is 100 frames). Their Openness is high because they keep low angular variation along their trajectories. On the other hand, their Conscientiousness dimension is lower than other dimensions because they keep low speeds in comparison to other groups.

The group OCEAN reflects the individuals OCEAN. An analysis of OCEAN in the country level is presented later in this section. Once we have the OCEAN values, individuals emotions are detected in the analyzed videos, as discussed in next section.

\subsection{Emotion analysis}
\label{sec:emotion_analysis}

\begin{figure}[h]
\centering
\includegraphics[scale=0.49]{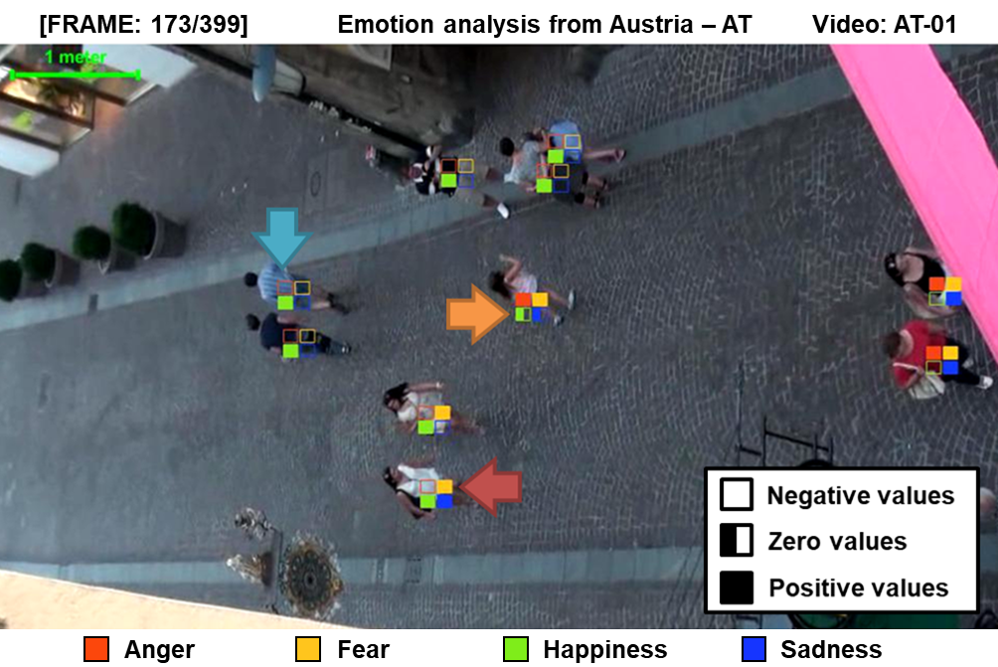}
\caption{Emotion representation based on the OCEAN mapping process. Each square represents one emotion, where a filled square represents that the person has a positive value from an emotion, a half filled square means that the emotion is neutral and a not filled square means that the person has a negative value from an emotion. Each color is related to one emotion: a red square is related to \textit{Anger}, an yellow square means \textit{Fear}, green indicates \textit{Happiness} and blue, \textit{Sadness}.}
\label{fig:emotion_rep}
\end{figure}

This section presents the emotion analysis in spontaneous videos. Figure~\ref{fig:emotion_rep} shows an example of the emotion detection in a video from Austria. A filled square represents that the person has a positive value from that emotion, a half filled square means that the emotion is neutral and a not filled square means that the person has a negative value from the emotion. For example, the highlighted person, with the blue arrow, got a negative value regarding the Anger emotion (the red square is not filled). It happens because this individual is interacting with the other (walking in the same direction and with similar angular variation), so collectivity and socialization levels are high and isolation level is low, consequently Anger receives a negative score.
 
Still in Figure~\ref{fig:emotion_rep}, the highlighted person with the orange arrow gets a zero value for the Happiness emotion (the green square is half filled). It happens because she/he is alone, changing orientation (angular variation), with high isolation. The person highlighted with the red arrow gets a positive value for the Fear (the yellow square is completely filled). It happens because this individual is moving slower and with high angular variation, so  his/her dimension C is low, generating a high value for Fear. The legend of colors is the following: red is related to Anger, yellow means Fear, green indicates Happiness and blue, Sadness. 

\begin{figure}[h]
\centering
\subfigure[fig:emo_0][Emotion of each individual in the scene]{\includegraphics[width=3.31in]{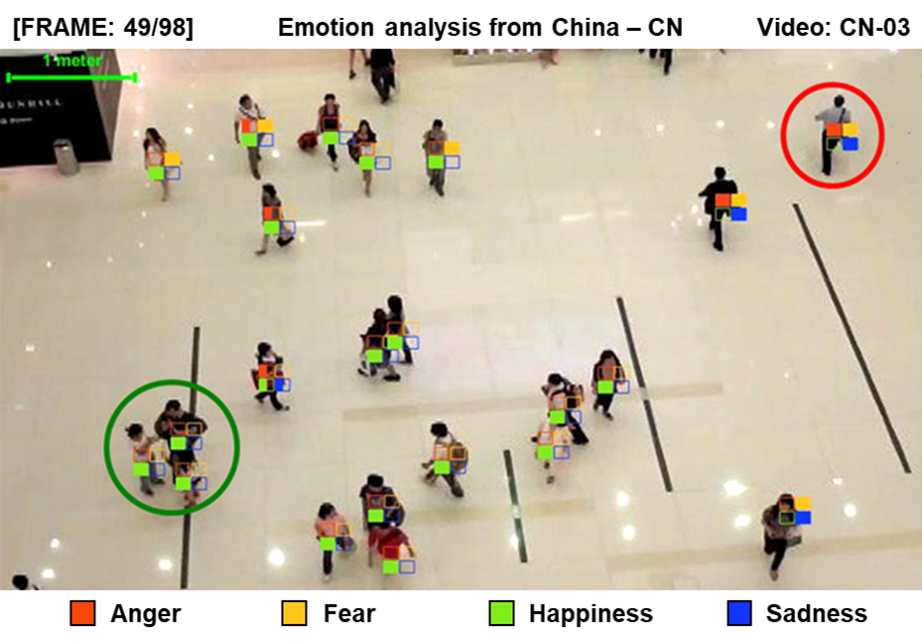}}
\subfigure[fig:emo_1][Group]{\includegraphics[width=1.67in]{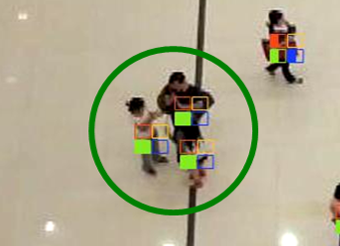}}
\subfigure[fig:emo_2][Individual]{\includegraphics[width=1.55in]{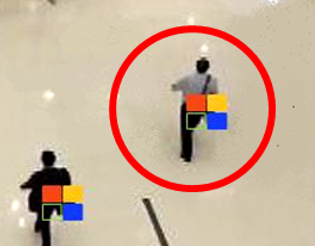}}
\caption{Emotion analysis in the video $CH-03$ from China: emotion of each individual in the scene (a), focused on a group (b) and one individual (c) highlighted. Based on our approach, individuals in the group (b) tend to be happy, while the individual alone (c) tends to experience negative emotions.}
\label{fig:emotion_example}
\end{figure}

In another example, showed in Figure~\ref{fig:emotion_example}(a), we highlight two different situations, a group (green circle) and an individual alone (red circle). It is interesting to notice that individuals who are part of a bigger group or have a high collectivity tend to be happy, as we can see in the highlighted group in Figure~\ref{fig:emotion_example}(b). On the other hand, individuals who are alone and distant from others tend to experience negative emotions (see an example in Figure~\ref{fig:emotion_example}(c)).

\subsection{OCEAN and Emotion Comparison with Literature}
\label{sec:compara}

In the next experiment we compare our results with available literature~\cite{costa07} about OCEAN values of different Countries. In the case of the present work we focus on Germany and Brazil, since in next section we also present a comparative about these two Countries. It is important to emphasize that the data registered in~\cite{costa07} was acquired based on subject answers on surveys and not based on their visual behavior in video sequences. We use OCEAN dimensions of the two analyzed Countries (Brazil and Germany) as presented by Costa et al.~\cite{costa07} as ground-truth in our approach. We evaluated our method in a set of $10$ available videos, where 2 videos were from Germany and 8 from Brazil.

\begin{figure}[h]
\centering
\subfigure[fig:BR_ALL][Brazil]{\includegraphics[width=3in]{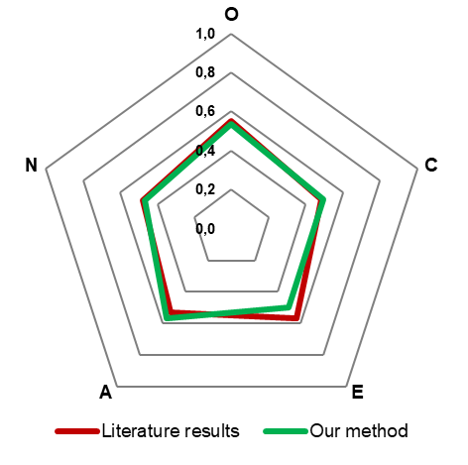}}
\subfigure[fig:GE_ALL][Germany]{\includegraphics[width=3in]{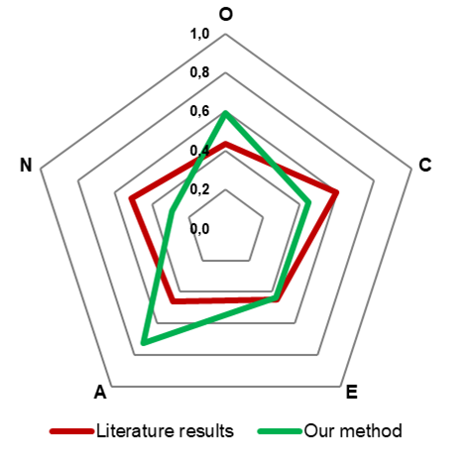}}
\caption{OCEAN comparison among our approach and literature~\cite{costa07} values: (a) data from Brazil and (b) data from Germany.}
\label{fig:all_countries}
\end{figure}

We evaluate the accuracy of our approach to detect the OCEAN values of the Countries based on percentage difference when compared with the literature results, considering all dimensions among all videos. Figure \ref{fig:all_countries} shows the results obtained with our approach for two countries in all OCEAN dimensions, in comparison with the literature~\cite{costa07}. It is interesting to highlight that results achieved for Brazil showed a higher accuracy, when compared to Germany. In the same time, this was the country with more videos to be analyzed.

\begin{figure}[h]
\centering
\includegraphics[width=3.3in]{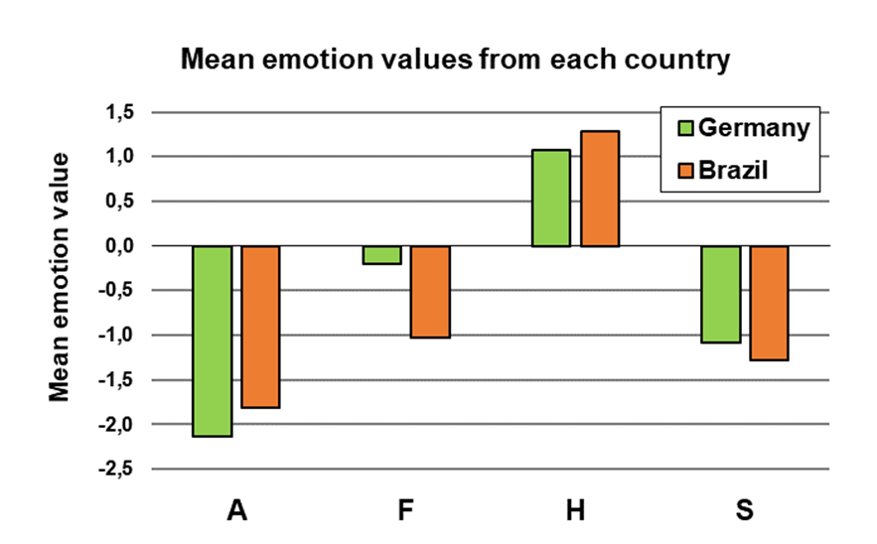}
\caption{Mean emotion from each country, considering all the 10 videos from the experiments (two from Germany and 8 from Brazil).}
\label{fig:mean_emotion_all}
\end{figure}

In the same way, we compute the emotions in the country level. For this, similar to the OCEAN approach, the emotion of each country is computed as the mean emotion from the videos from that country. Figure~\ref{fig:mean_emotion_all} shows the mean emotion from the analyzed countries in this experiment. As far as we know, there is no data to be compared in this matter.

\subsection{OCEAN and emotion analysis in the Fundamental Diagram experiment}
\label{sec:ocean_emotion_fd}

\begin{figure}[h]
\centering
\includegraphics[width=3.3in]{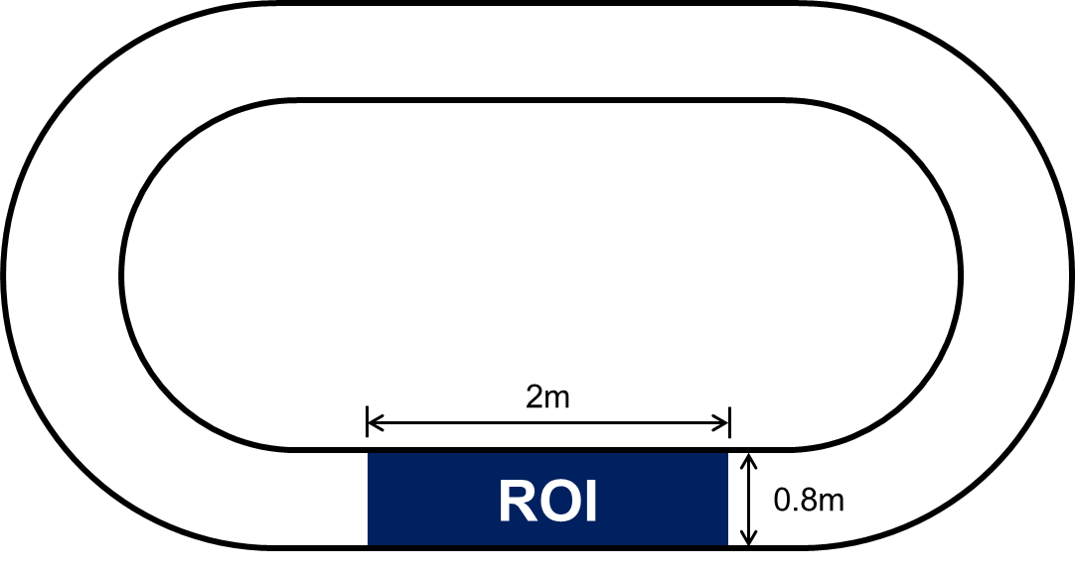}
\caption{Sketch of experimental setup according to \cite{Chattaraj:2009}. The Region Of Interest (ROI) is a rectangle of 2 x 0.8 meters.}
\label{fig:config_exp}
\end{figure}

In last sections we were interested about detecting personality and emotion traits in videos from different Countries, in public spaces. However, since the contexts are not the same, e.g. people can be close or faraway from others not just because their personality but because the contexts they are evolving on or the relationship with spaces are not the same. Consequently, this research presents such difficult challenge to solve in order to evaluate/validate results. The ideal is to discard the spacial context in order to analyze only the people behavior, from different Countries, while executing the same task. 
Therefore, we found the Fundamental Diagram applied to pedestrians, as proposed in~\cite{Chattaraj:2009}. We detect personality and emotion traits in video sequences from two countries (Brazil and Germany~\footnote{We have access to such videos thanks to the authors of database of PED experiments, available at \url{http://ped.fz-juelich.de/db/}.}), where all the individuals are performing exactly the same task. i.e. walking in a controlled space. The experiments have been performed in an environment setup as illustrated in Figure~\ref{fig:config_exp}.

This experiment was conducted as described in \cite{Chattaraj:2009}, being the same in all the two countries, with the same populations ($N=15$, $25$ and $34$). The corridor was built up with markers and tape on the ground. Its size and shape is presented in Figure~\ref{fig:config_exp}. The length of the corridor is $17.3m$, while the width of the passageway is $0.8m$, which is sufficient for a single person walk. In addition, we can observe on the bottom of Figure~\ref{fig:config_exp} a rectangle of 2 x 0.8 meters which illustrates the Region of Interest (ROI) where the populations were captured to be analyzed, as proposed in~\cite{Chattaraj:2009}.

\begin{figure*}[t]
\centering
\subfigure[fig:fd_bra15][$N=15$ (Brazil)]{\includegraphics[width=8cm,height=5.2cm]{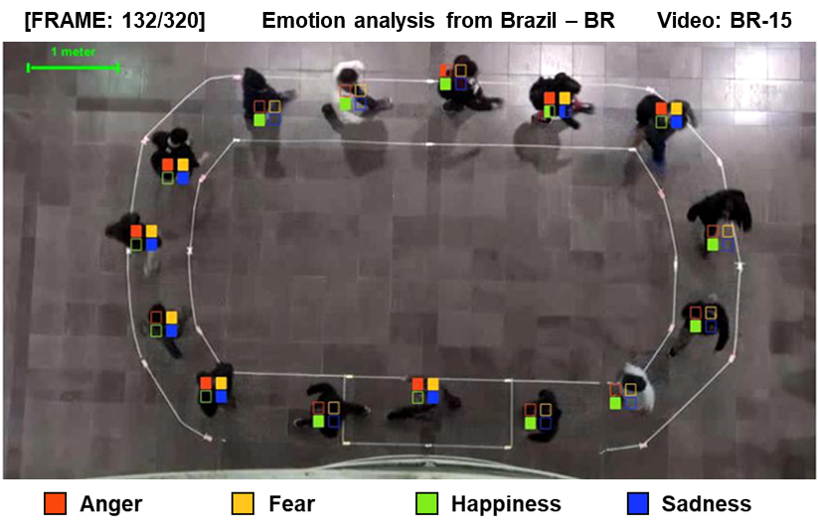}} \qquad
\subfigure[fig:fd_ger15][$N=15$ (Germany)]{\includegraphics[width=8cm,height=5.2cm]{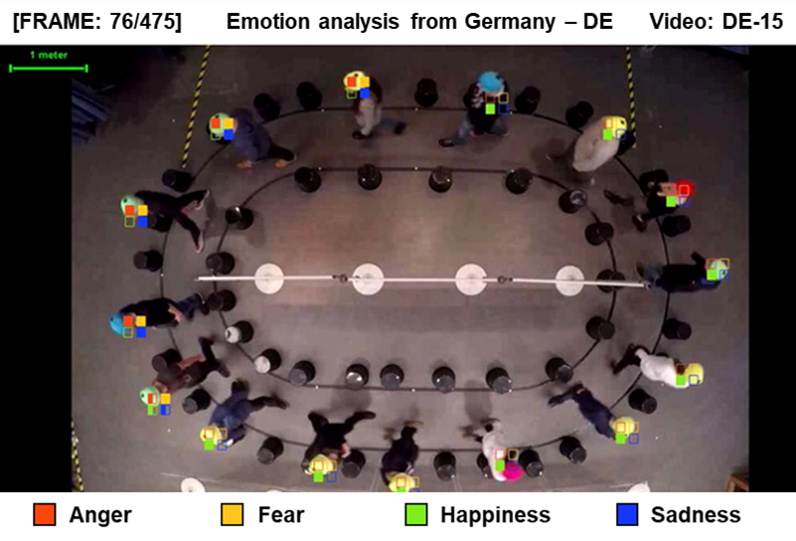}}
\subfigure[fig:fd_bra25][$N=25$ (Brazil)]{\includegraphics[width=8cm,height=5.2cm]{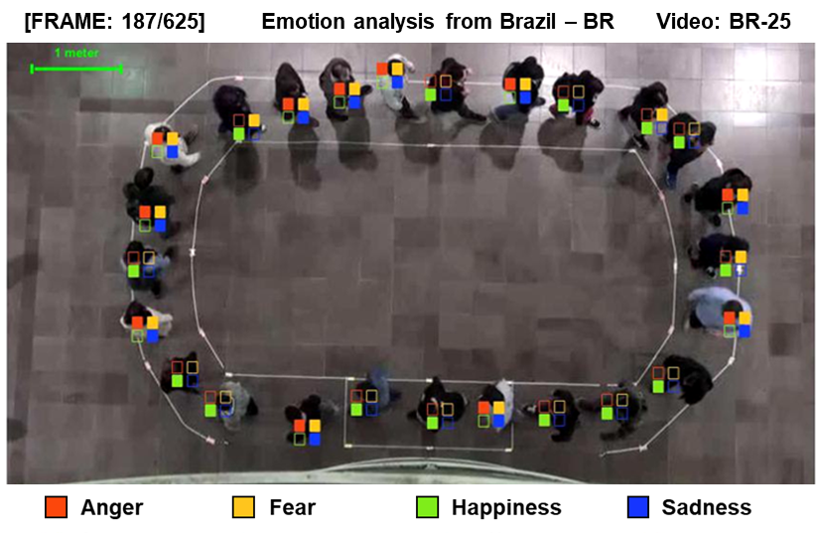}} \qquad
\subfigure[fig:fd_ger25][$N=25$ (Germany)]{\includegraphics[width=8cm,height=5.2cm]{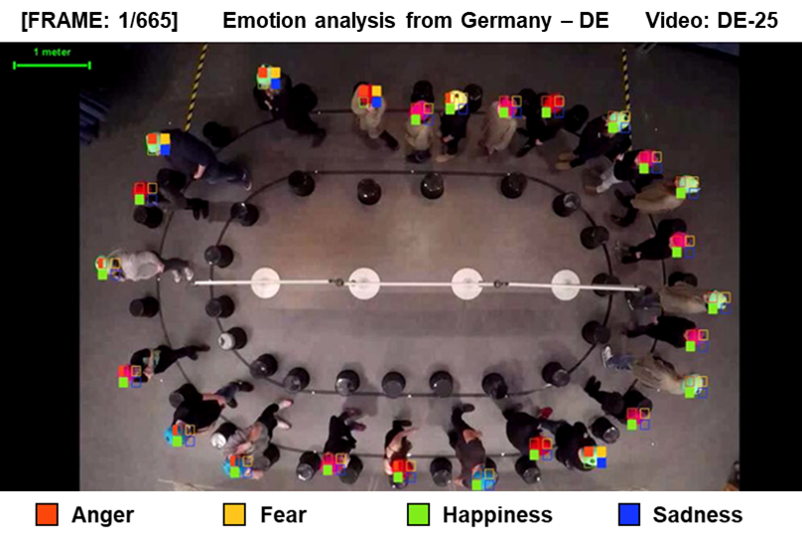}}
\subfigure[fig:fd_bra34][$N=34$ (Brazil)]{\includegraphics[width=8cm,height=5.2cm]{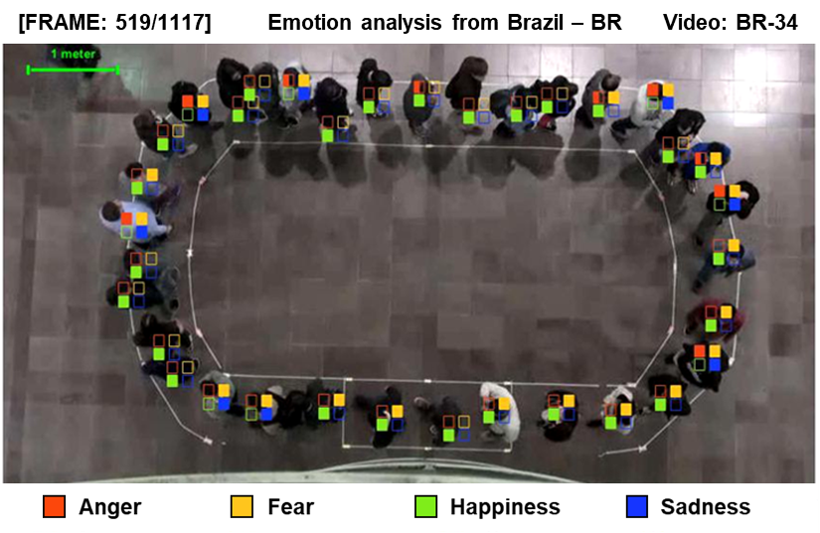}} \qquad
\subfigure[fig:fd_ger34][$N=34$ (Germany)]{\includegraphics[width=8cm,height=5.2cm]{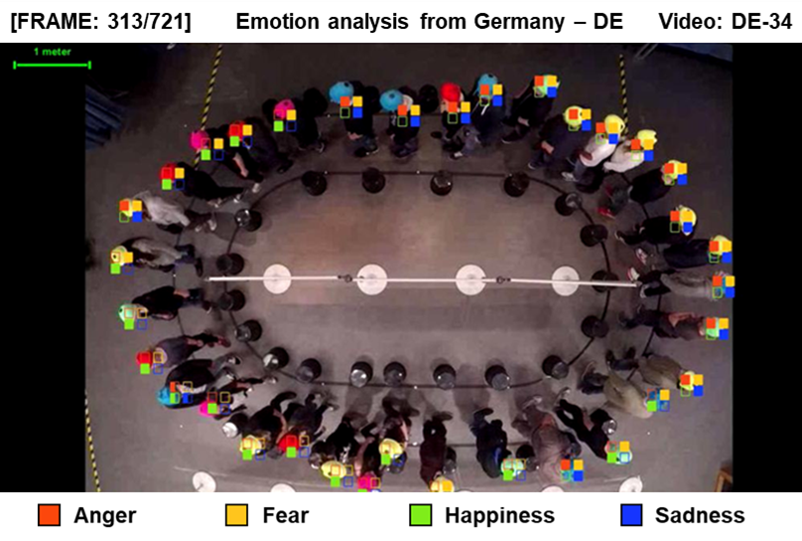}}
\caption{Emotion analysis from the individuals performing the experiment. \textbf{Left side}: experiment from Brazil, with 15 (a), 25 (c) and 34 individuals (e). \textbf{Right side}: experiment from Germany, with 15 (b), 25 (d) and 34 individuals (f).}
\label{fig:emotion_fd}
\end{figure*}

Figure~\ref{fig:emotion_fd} shows examples of emotion detection performed by individuals in the experiment of Fundamental Diagram. Pictures show the performed tests with different sizes of population in both Countries.

\begin{figure}[h]
\centering
\subfigure[fig:N_15][$N=15$]{\includegraphics[width=3in]{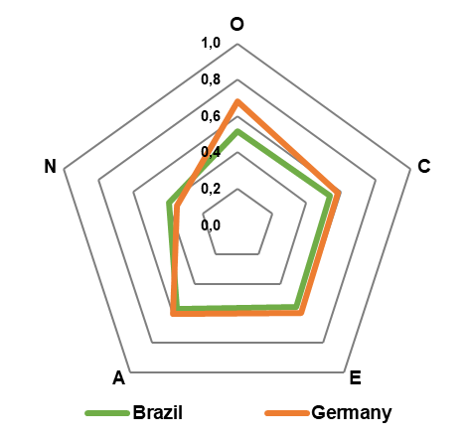}}
\subfigure[fig:N_34][$N=34$]{\includegraphics[width=3in]{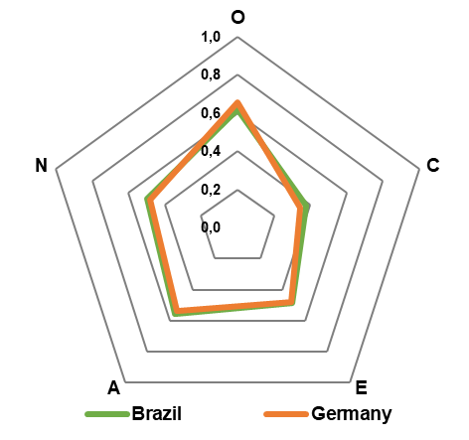}}
\caption{Averaged OCEAN values observed in each country: (a) when $N=15$ individuals and (b) when $N=34$ individuals.}
\label{fig:ocean_fd}
\end{figure}

Figure~\ref{fig:ocean_fd} shows the OCEAN values from Brazil and Germany in the experiments with extreme sizes of populations, i.e. $N=15$ and $N=34$. Based only on a visual inspection, we can easily perceive that OCEAN values from the both Countries are more similar when the density of people is higher in the experiment. It can indicate that people assumes group-level behavior instead of individual-level behavior caused by the higher density and the lack of free space. It agrees with several theories about mass behavior as discussed in~\cite{vilanova2017} and ~\cite{LE_BON_THE_CROWD}.

\begin{figure}[h]
\centering
\includegraphics[width=3.3in]{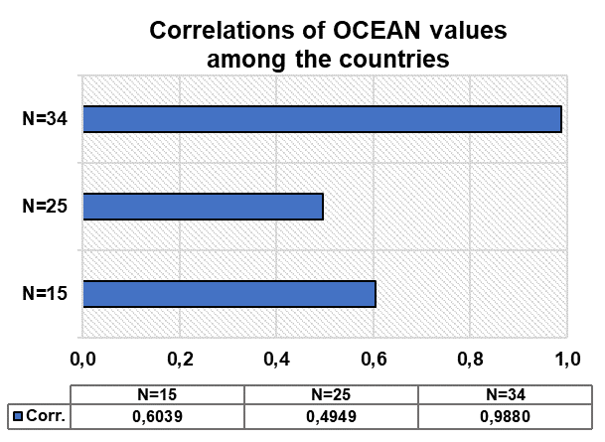}
\caption{Pearson's correlation of OCEAN values obtained in videos from Germany and Brazil, with different populations ($N=15$, $N=25$ and $N=34$).}
\label{fig:corr_ocean}
\end{figure}

\begin{figure}[h]
\centering
\includegraphics[width=3.3in]{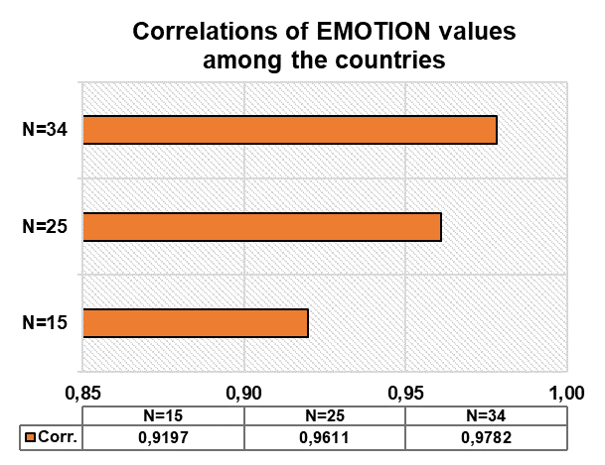}
\caption{Pearson's correlation of emotion values obtained in videos from Germany and Brazil, with different populations ($N=15$, $N=25$ and $N=34$).}
\label{fig:corr_emotion}
\end{figure}

In this analysis, we also compute Pearson's correlation to find out the similarity between the two countries in both OCEAN and emotion aspects. Figure~\ref{fig:corr_ocean} shows the Pearson's correlation of OCEAN values among the countries, while Figure~\ref{fig:corr_emotion} shows the Pearson's correlation of the emotion values.


\begin{figure}[h]
\centering
\subfigure[fig:GER_emo][Germany]{\includegraphics[width=3.3in]{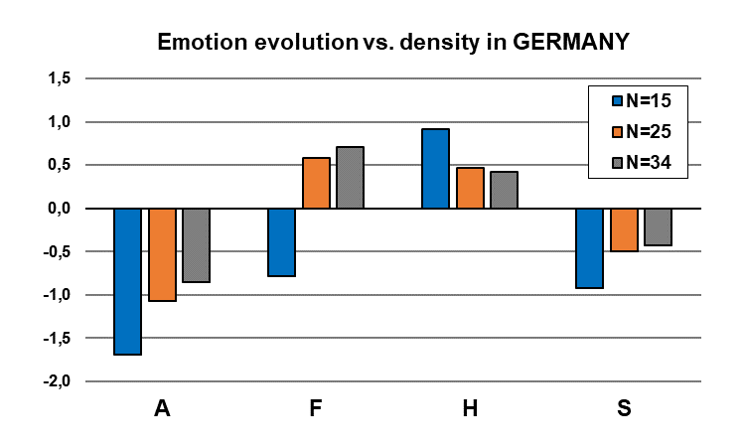}}
\subfigure[fig:BRA_emo][Brazil]{\includegraphics[width=3.3in]{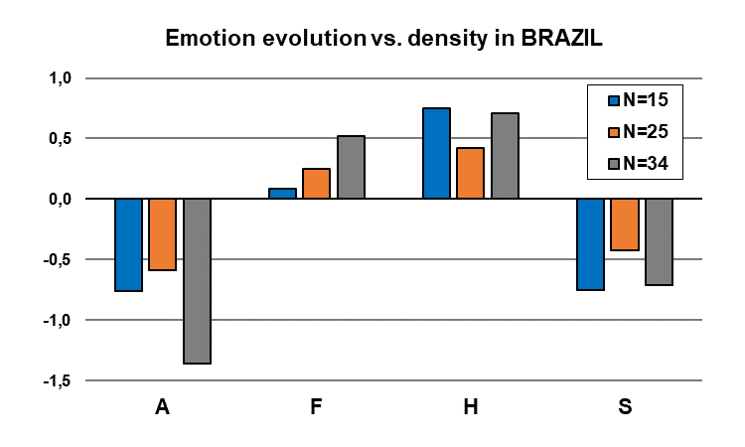}}
\caption{Impact of emotion on the variation of densities ($N=15$, $N=25$ and $N=34$) in both countries: (a) Germany and (b) Brazil.}
\label{fig:emotion_density}
\end{figure}

Considering the both Figures~\ref{fig:corr_ocean} and \ref{fig:corr_emotion}, it is possible to see that, except from the population $n=25$ in OCEAN experiment, as the density of people in the experiment increases, the correlation among the countries also increases.
We also included an investigation regarding the emotion and OCEAN impacted by the density of people. This was possible because the performed task is the same (i.e. individuals are walking in the same predefined environment) while only the number of people increases. Figure~\ref{fig:emotion_density}(a) shows how the emotion values vary according to the density of people in videos from Germany. It is interesting to see how the emotions Anger, Fear and Sadness increases proportionally as the density increases too. On the other hand, Happiness emotion decreases proportionally as a function of observed density. Indeed, the data observed in Germany was in accordance to what was empirically expected in our hypothesis, i.e. the only positive emotion (H) decreases as the density increases. However, the computed emotion for Brazil was not so well behaved as in Germany, and certainly it could be better investigated in a future work. One possible explanation is that Brazilian people were colleagues/friends while in Germany, they were related in~\cite{Chattaraj:2009} as volunteers to the experiment.

Although we did not find any information about emotion and personality detection in videos from different Countries related in literature, these results seem promising in order to understand people personality, emotion and cultural differences in video sequences. 

\subsection{Quantitative Analysis: Personal Space}
\label{sec:personal_space}

\begin{figure}[h]
\centering
\includegraphics[width=3.3in]{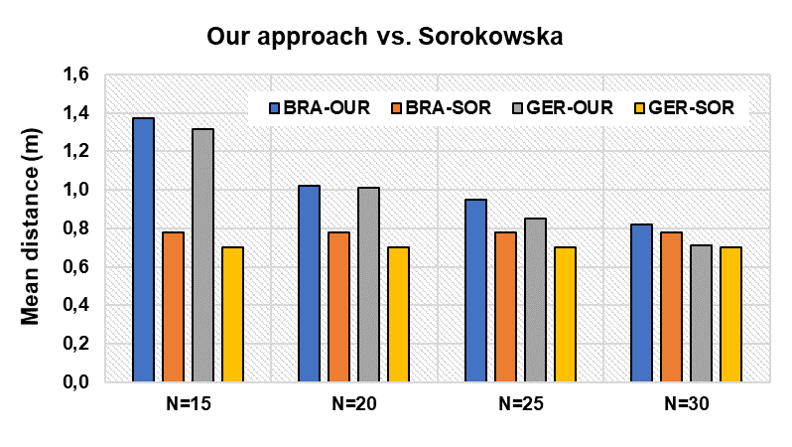}
\caption{Preferred distances observed in our approach versus Sorokowska~\cite{sorokowska:2017} in different populations ($N=15$, $N=20$, $N=25$ and $N=30$).}
\label{fig:our_vs_soro}
\end{figure}

In this experiment we performed a comparison among the preferred distance people keep from others, as evaluated in a study performed by \cite{sorokowska:2017}. Results obtained with the experiment performed in our approach are compared with results of \cite{sorokowska:2017}, as shown in Figure~\ref{fig:our_vs_soro}. In the Sorokowska work, the answers were given on a distance (0-220 cm) scale anchored by two human-like figures, labeled A and B. Participants were asked to imagine that he or she is Person A. The participant was asked to rate how close a Person B could approach, so that he or she would feel comfortable in a conversation with Person B.

In our approach we measure the distances a person A keeps from a person B right in front of he or she, in the video sequences. For the comparison, in our approach, we use the distances from the Fundamental Diagram video sequences varying the people number ($N=15$, $N=20$, $N=25$ and $N=30$) and from the Sorokowska's approach we select the evaluation from acquaintance people, where the people are not close neither strangers, similar to people in our experiment. As we can see in Figure~\ref{fig:our_vs_soro}, in spite of the fact that distances from our approach are higher than the ones from Sorokowska, when there are few people ($N=15$ and $N=20$), the proportion is similar in both scenarios. In addition, in the experiment $N=30$, the distances observed in our approach are close to those obtained in the Sorokowska study. In all scenarios, people from Brazil keeps higher distances from others than people from Germany. The same happens in Sorokowska's research. Although they are different experiments (videos and surveys), our method indicates that there is a correlation between people feeling in abstracted environment (as in surveys) and as they behave in video sequences.

\section{Final considerations}
\label{sec:conclusions}

In this paper, we describe a way to detect individual-level traits of emotion and personality observed in video sequences, based on individuals and groups features. We propose to detect OCEAN personality traits and compared with data from different countries existent in the literature. In addition, based on OCEAN we compute 4 traits of emotion defined in OCC model. We believe the results are promising and video sequences can be used to detect personality and emotion, what can help us to understand people behavior in video sequences.

An important challenge in this area is the comparison with real life data. In this work we successfully compared our results with OCEAN from two specific countries (Brazil and Germany) present in Psychological literature. As one particular aspect to be considered in behavior analysis is the context and environment in which individuals behave, we decided to exclude such variations by fixing tasks that the tested populations were required to execute. This is why we used Fundamental Diagram for pedestrians. To do that we performed a full experiment in Brazil to serve as benchmark to our research. Therefore, we measured the preferred personal distance from individuals in FD in order to present a quantitative data analysis.

The results obtained from our approach indicate that our model generates coherent information when compared to data provided in available literature, as shown in various analysis. It is important to note that the mapping to OCEAN and EMOTION dimensions was empirically defined through equations using data extracted from computer vision. NEO PI-R results in the literature measured these dimensions by considering a different type of information (subjective responses of individuals collected through questionnaires). The results of preferred distances presented by Sorokowska et al.~\cite{sorokowska:2017} also were based on subjective responses informed by individuals.

For our future work, we consider to select items related to collectiviness from the International Personality Item Pool (http://ipip.ori.org/), which originated the NEO PI-R, to establish a questionnaire that can identify individual-level traits that correspond to group behaviors and could be analyzed computationally. One way to do this would be to ask participants to answer the proposed questionnaire, then divide these participants into groups with high and low scores, and, finally, ask these groups to perform some behavioral tasks; tasks that could be mapped in videos of crowds. Such a set of individual-level traits related to group behavior will not only increase the accuracy of group analysis tools by means of videos as well as group simulations based on the unique characteristics of each agent.


\bibliographystyle{spmpsci}      

\bibliography{refs}

\end{document}